\documentclass[10pt,twocolumn,letterpaper]{article}
\usepackage[pagenumbers]{cvpr} 

\usepackage{overpic}
\usepackage{graphicx}
\usepackage{amsmath}
\usepackage{amssymb}
\usepackage{booktabs}

\def\etc{{\em etc}}

\usepackage[pagebackref,breaklinks,colorlinks]{hyperref}

\usepackage[capitalize]{cleveref}
\crefname{section}{Sec.}{Secs.}
\Crefname{section}{Section}{Sections}
\Crefname{table}{Table}{Tables}
\crefname{table}{Tab.}{Tabs.}

\begin{document}

\title{Long Range Pooling for 3D Large-Scale Scene Understanding}

\author{Xiang-Li Li\textsuperscript{\rm 1} ~~ 
Meng-Hao Guo\textsuperscript{\rm 1} ~~
Tai-Jiang Mu\textsuperscript{\rm 1} ~~
Ralph R. Martin\textsuperscript{\rm 2} ~~
Shi-Min Hu\textsuperscript{\rm 1}
\\
\textsuperscript{\rm 1}Tsinghua University ~~ 
\textsuperscript{\rm 2}Cardiff University \\ 
{\tt\small lixl19@mails.tsinghua.edu.cn, gmh20@mails.tsinghua.edu.cn}\\
{\tt\small taijiang@tsinghua.edu.cn, martinrr@cardiff.ac.uk,  shimin@tsinghua.edu.cn}
}
\maketitle

\begin{abstract}
Inspired by the success of recent vision transformers and large kernel design in convolutional neural networks (CNNs), in this paper, we analyze and explore essential reasons for their success.
We claim two factors that are critical for 3D large-scale scene understanding: \textbf{a larger receptive field} and \textbf{operations with greater non-linearity}.
The former is responsible for providing 
long range contexts and the latter can enhance the capacity of the network.
To achieve the above properties, we propose a simple yet effective long range pooling (LRP) module using dilation max pooling, which provides a network with a large adaptive receptive field.
LRP has few parameters, and can be readily added to current CNNs.
Also, based on LRP, we present an entire network architecture, LRPNet, for 3D understanding.
Ablation studies are presented to support our claims, 
and show that the LRP module achieves better results than 
large kernel convolution yet with reduced computation, due to its non-linearity. 
We also demonstrate the superiority of LRPNet on various benchmarks: LRPNet performs the best on ScanNet and surpasses other CNN-based methods on S3DIS and Matterport3D.
Code will be made publicly available.

\end{abstract}

\section{Introduction}
\label{sec:intro}

With the rapid development of 3D sensors, 
more and more 3D data is becoming available from a variety of applications such as autonomous driving, robotics, and augmented/virtual reality. This 3D data  
requires analyzing and understanding.
Efficient and effective processing of such 3D data 
has become an important challenge.

Various data structures, including point clouds, meshes, multi-view images, voxels, \etc,
have been proposed for representing 3D data~\cite{xiao2020survey},
and different network architectures are designed to process them.
Unlike 2D image data, a point cloud or mesh,
is irregular and unordered.
These characteristics mean that typical CNNs 
cannot directly be applied to such 3D data.
Thus, specially designed networks such as 
MLPs~\cite{qi2017pointnet}, 
CNNs~\cite{li2018pointcnn,wu2019pointconv,hu2022subdivision}, 
GNNs~\cite{schult2020dualconvmesh,wang2019dynamic} and 
transformers~\cite{guo2021pct,zhao2021point,mao2021voxel} 
are proposed to perform effective deep learning on them. 
However, processing large-scale point cloud or mesh
is computationally expensive.
Multi-view images contain
multiple different views of a scene.
Typically, CNNs are used to process each view 
independently and a fusion module is used to combine the results.
Nevertheless, there is unavoidable 3D information loss due to the finite or insufficient number of views. 
Voxel data has the benefit of regularity like images, 
even if the on-surface voxels only contain sparse data.
The simplicity of the structure helps to maintain high performance, so this paper focuses on processing voxel data.

\begin{figure}
    \centering
    \includegraphics[width=\columnwidth]{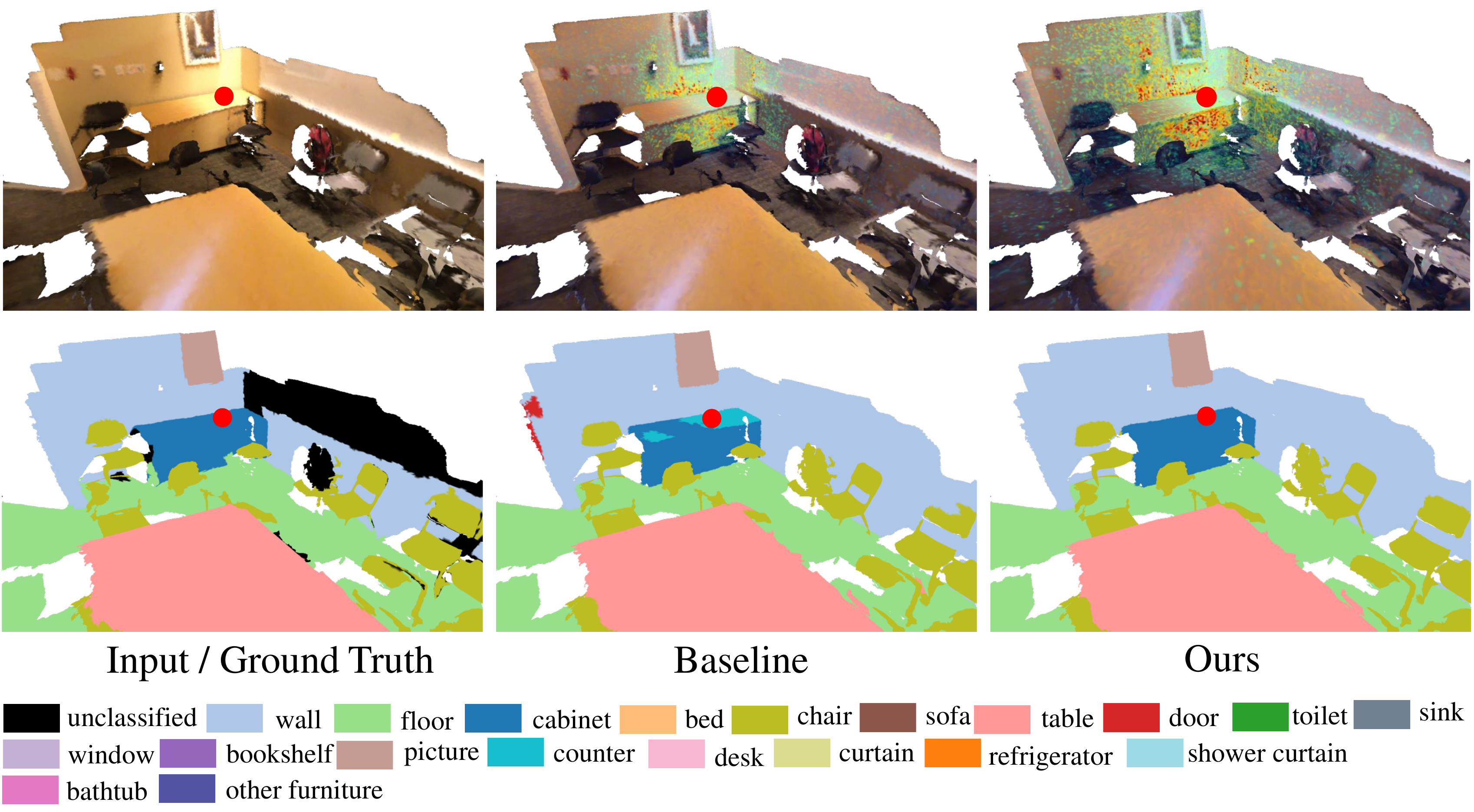}
    \caption{Qualitative results of Effective Receptive Field (ERF)~\cite{luo2016understanding}. \textbf{Left}: the input and ground truth. \textbf{Middle}: the ERF and results of baseline described in section~\ref{sec:network}.
    \textbf{Right}: the ERF and results of LRPNet (ours). The stained areas around the positions of interest (red dots) represent the range of receptive fields, with green to red representing the increasing strength of response. Our method correctly segments the cabinet from the wall with the proposed effective receptive field.}
    \label{fig:heatmap_compare}
\end{figure}

Learning on 3D voxels can be easily implemented by directly extending the well studied 2D CNN networks to 3D~\cite{wu20153d,riegler2017octnet,wang2017cnn}.
Considering that 3D voxel data is inherently sparse, some works usually adopt specially designed 3D sparse CNNs~\cite{choy20194d, graham20183d, shi2020pv} for large-scale scene understanding. 
Since only the surface of the real scene data has values, the neighbors around each voxel do not necessarily belong to the same object, so a large receptive field is more conducive to feature extraction.
Sparse convolution has advantages in modeling local structures, 
but it ignores long range contexts,
which are critical for 3D scene understanding tasks. 
Transformers~\cite{wu2022point,wu2022pointconvformer,lai2022stratified,guo2021pct} 
have proved useful in processing 3D scenes, as they can capture the global relationship with their
larger receptive field and a better way to interact. However, they usually have a quadratic computational complexity, which brings a heavy computing cost.

A straightforward way to incorporate the long range contexts into the learning of 3D voxel data is to exploit a large kernel 3D convolution.
However, the number of network parameters and the amount of 
computation would increase cubically due to the additional dimension compared to 2D images.
Besides, current ways of feature interaction or aggregation, such as average pooling and convolution, usually adopt a linear combination of features of all the locations in the receptive field. 
This works for 2D images since all the locations in the receptive field have valid values, which, however, does not hold for 3D voxels due to the sparsity nature of 3D scenes.
Directly applying such linear interaction or aggregation on the voxel that has few neighbors in the receptive field would make the feature of that voxel too small or over-smoothed and thus less informative.

Taking the above into consideration, and to achieve a trade-off between the quality of results and computation, we propose long range pooling (LRP), a simple yet effective 
module for 3D scene segmentation.
Compared with previous sparse convolution networks~\cite{choy20194d,graham20183d},
LRP is capable of increasing the effective receptive field with a negligible amount of computational overhead.
Specifically, we achieve a large receptive field by proposing a novel dilation max pooling to enhance the non-linearity of the neural network. 
We further add a receptive field selection module, so that each voxel can choose a suitable receptive field, which is adaptive to the distribution of voxels. The above two components comprise the LRP module.
Unlike dilation convolution, which is often used to enlarge the receptive field for 2D images~\cite{guo2022segnext,guo2022visual}, our method can achieve a large receptive field with fewer parameters and computation by using dilation max pooling. 
Furthermore, LRP is a simple, efficient and parameterless module that can be readily incorporated into other networks. 
We construct a more capable neural network, \emph{LRPNet}, by adding LRP at the end of each stage of the sparse convolution network, introduced by VMNet~\cite{hu2021vmnet}.

Experimental results show that LRPNet achieves a significant improvement 
in 3D segmentation accuracy on large-scale scene datasets, including ScanNet~\cite{dai2017scannet}, S3DIS~\cite{armeni20163d} and Matterport3D~\cite{chang2017matterport3d}.
Qualitative results are illustrated in Figure~\ref{fig:heatmap_compare}, which shows the improvement of a larger receptive field.
We also experimentally compare the effects of the different receptive fields by reducing the number of dilation max pooling of LRP.
Moreover, we explore the influence of non-linearity of LRP module by replacing max pooling with average pooling or convolution. 
Ablation results show that a larger receptive field and operations with greater non-linearity will improve the segmentation accuracy.

Our contributions are thus:
\begin{itemize}
    \item a simple and effective module, the long range pooling (LRP) module, which provides a network with a 
    large adaptive receptive field without a large number of parameters, 
    \item a demonstration that a larger receptive field and operations with greater non-linearity enhance the capacity of a sparse convolution network, and
    \item a simple sparse convolution network using the LRP module, which achieves superior 3D segmentation results on various large-scale 3D scene benchmarks.
    
\end{itemize}

\section{Related Work}
\label{sec:related}

\subsection{Deep Learning in 3D Vision}
Deep learning has achieved great success in the field of 3D vision in recent years~\cite{xiao2020survey}.
However, compared to 2D images, 3D data is complex and has diverse representations, such as multi-view images, point clouds, voxels, meshes and so on. 
With the availability of some large 3D datasets~\cite{Geiger2012CVPR, wu20153d,armeni20163d,dai2017scannet}, 3D vision has entered a period of rapid development.
Due to the success of neural networks in processing 2D images, some works~\cite{su2015multi,shi2015deeppano,sinha2016deep} have used 2D convolutional networks to extract features from 2D views, and applied a fusion module for 3D understanding. In fact, the raw data is usually in the form of a  point cloud, so some researchers~\cite{qi2017pointnet,qi2017pointnet++,li2018pointcnn,thomas2019kpconv,hu2020randla,wu2019pointconv} have designed point neural networks for point cloud processing.
Due to the high computational complexity of processing point cloud networks, 
researchers~\cite{choy20194d, graham20183d, chen2022scaling} began to use hash functions to search for neighbors, and to apply 3D sparse convolution to achieve efficient voxel understanding.
Meshes have not only position information, but also topological structure. Specially designed mesh neural networks~\cite{smirnov2021hodgenet,hu2022subdivision,hanocka2019meshcnn,lahav2020meshwalker,sharp2022diffusionnet} take advantage of this topological information. 
In addition, to further improve the effectiveness of 3D semantic understanding, fusion learning of multiple types of 3D data has also gradually developed, such as point cloud + mesh~\cite{schult2020dualconvmesh}, point cloud + image~\cite{hu2021bidirectional}, voxel + mesh~\cite{hu2021vmnet}, and so on.

Most works~\cite{qi2017pointnet,hanocka2019meshcnn,choy20194d} focus on how neural networks can make better use of 3D data, and thus construct specific modules to enhance the ability to extract features. Point networks from PointNet~\cite{qi2017pointnet} to PointConv~\cite{wu2019pointconv}, PointCNN~\cite{li2018pointcnn} improve the ability to extract local information. Going from point networks~\cite{qi2017pointnet++,li2018pointcnn} to point transformers~\cite{guo2021pct,wu2022pointconvformer,wu2022point,lai2022stratified} expands the scope of the model's receptive field and the ability to recognize the interaction between features. However, the computational demands of these methods are higher, hindering their application to large scenes. In order to balance computational efficiency and accuracy, voxel-based networks~\cite{choy20194d,graham20183d} use 3D sparse convolution to aggregate local features. 
Limited by a large number of 3D convolution parameters, it is difficult to improve results by directly increasing the receptive field. 
LargeKernel3D~\cite{chen2022scaling} expands the receptive field of 3D convolution by depth-wise convolution and dilation convolution. 
However, compared to transformers~\cite{guo2021pct,wu2022pointconvformer,wu2022point,lai2022stratified}, the receptive field of this approach is still small, and it introduces more parameters and computation. To solve these problems,  we propose the long range pooling method, which can expand the receptive field with few additional parameters and introduce operations with greater non-linearity to enhance the capability of the neural network.

\subsection{Vision Transformers}
\label{sec:related_vit}

The transformer  network architecture 
comes from natural language processing~\cite{vaswani2017attention,devlin2018bert}.
Recently, due to its strong modeling capability,
it has quickly provided leading methods for various vision tasks, including 
image classification~\cite{dosovitskiy2020image, guo2022beyond},
object detection~\cite{liu2021swin,li2022exploring},
semantic segmentation~\cite{zheng2021rethinking,xie2021segformer},
image generation~\cite{jiang2021transgan,lee2021vitgan},
and self-supervised learning~\cite{he2022masked,bao2021beit}---see the surveys in~\cite{khan2022transformers,guo2022attention}.
The transformer architecture is also introduced into 3D vision by PCT~\cite{guo2021pct} and PT~\cite{zhao2021point} almost simultaneously, 
which propose a 3D transformer 
based on global attention and 
a 3D transformer based on local attention, respectively.

The core module of a transformer is the self-attention block, 
which models relationships by calculating pairwise similarity
between any two feature points.
We believe the success of self-attention arises for two reasons: 
(i) self-attention captures
\emph{long range dependencies}, and (ii) the matrix multiplication of attention and value, and softmax function  provide \emph{strong non-linearity}.

\subsection{Large Kernel Design in CNNs}
\label{sec:related_lk}
Inspired by the success of vision transformers, 
researchers have challenged the traditional small kernel design of CNNs~\cite{simonyan2014very,he2016deep} 
and suggested the use of large convolution kernels 
for visual tasks~\cite{liu2022convnet, guo2022visual,ding2022scaling,liu2022more,yang2022focal,guo2022segnext,rao2022hornet}.
For example, ConvNeXt~\cite{liu2022convnet} suggest directly adopting a 7$\times$7 depth-wise convolution, while the 
Visual Attention Network (VAN)~\cite{guo2022visual} uses a kernel size of 21 $\times$ 21 
 and introduces an attention mechanism. 
RepLKNet~\cite{ding2022scaling} introduces a 31 $\times$ 31 convolution by
using a reparameterization technique.
Recently, this design is also introduced into
3D field by LargeKernel3D~\cite{chen2022scaling}.
Analyzing these previous works, 
we observe that VAN~\cite{guo2022visual} gives the best results, which we believe is because
it introduces non-linearity via the Hadamard product, in addition to long range dependencies.

\section{Method}
\label{sec:method}
Our analysis above suggests that long range interactions and greater non-linearity may be the key to success.
Accordingly, we have designed a simple yet effective long range pooling (LRP) module, based on these principles.  
We now introduce our LRP and LRPNet in detail.

\subsection{Dilation Module for 3D Voxel} 
\label{sec:dilated_pool}
To our knowledge, we are the first to implement a large receptive field network for processing 3D voxel by using dilation max pooling. 
In order to make the paper self-contained,
we revisit the decomposition of a large kernel by dilated operations in 2D CNNs. 

In fact, general convolutional neural networks can implicitly realize large receptive fields over the whole network. 
With increasing network depth, the receptive field of the last layer of features gradually increases---this is one of the reasons why deep neural networks can be effective.
Meanwhile, previous works~\cite{guo2022segnext,guo2022attention} have shown that increasing the local receptive field can give better results than increasing the depth.
On the other hand, it is almost impossible to apply large kernels directly to the network because of their high computation load and large number of parameters.
Currently, the commonly used large kernel size is $31\times 31$ in 2D images~\cite{ding2022scaling}, which improves accuracy while also introducing more parameters and computation.
Therefore, most of the previous works have decomposed the large kernel module and used the computationally friendly small kernel module to approximate the large kernel convolution~\cite{guo2022visual}. 

For 2D images, several dilation convolutions are usually used to achieve a decomposed large kernel convolution.
However, 3D data is sparse, and the sparsity affects the response strength of the convolution, 
which makes large kernel decomposition more difficult. 
Besides, due to the sparsity, the weight convergence of convolution is too slow. 
Therefore, we use dilated pooling to achieve a large receptive field while keeping a low computational cost. 

\subsection{Long Range Pooling Module} 
\label{sec:lrp}
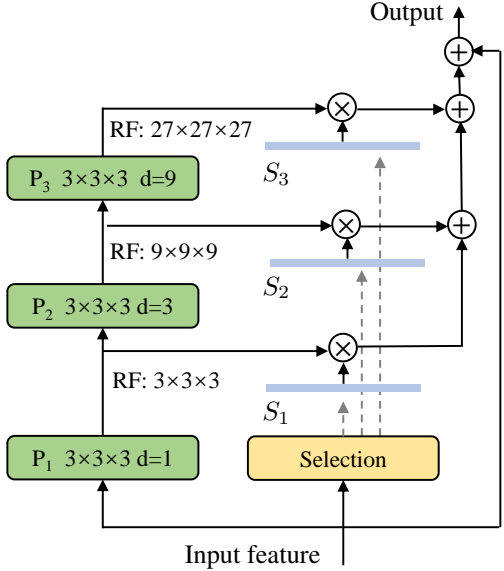
\begin{figure}[t]
    \centering
    \begin{overpic}[width=0.86\columnwidth]{ 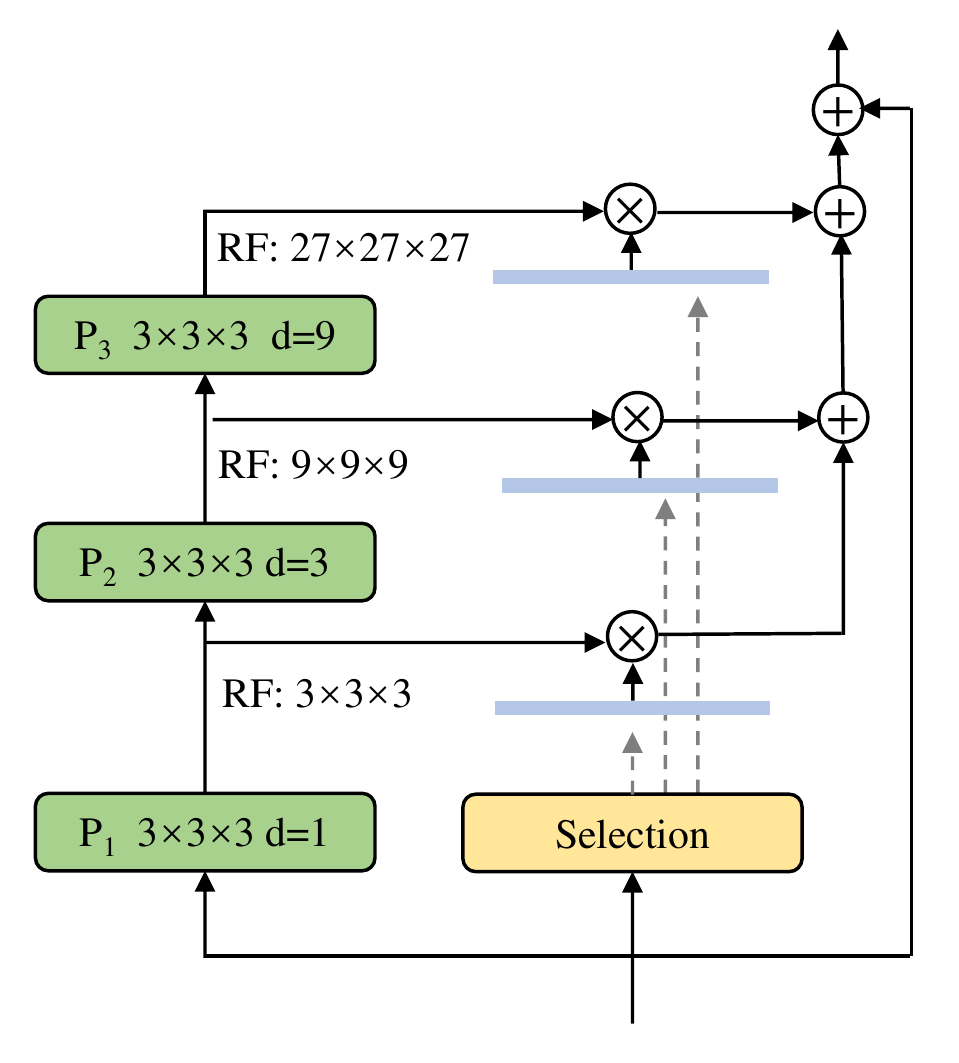}
        \put(33,4){Input feature}
        \put(46,28){$S_1$}
        \put(46,49){$S_2$}
        \put(46,68){$S_3$}
        \put(64, 95){Output}
    \end{overpic}
    \caption{Long Range Pooling Module. Dilation max pooling is used to enlarge the receptive field and also outputs features of the different receptive field. \textbf{$P_1, P_2, P_3$}: max pooling with kernel size $3\times 3 \times 3$ and dilation $1, 3, 9$. \textbf{RF}: the receptive field. \textbf{Selection}: a linear module producing the selection weight for each voxel.}
    \label{fig:lrp_module}
\end{figure}

Two direct strategies can achieve a
large receptive field and non-linearity: self-attention, and max pooling with a large window size. 
The large computational load placed by self-attention limits its application to 3D data. 
Max pooling also imposes huge computational demands as the window grows.
We avoid the computational cost of max pooling by introducing sparse dilated pooling.

As shown in Figure~\ref{fig:lrp_module},
we approximate pooling with a large window size
 by stacking three 3$\times$3$\times$3 
pooling modules with different dilation rates. 
This achieves a progressive increase in receptive field size while keeping the computational cost low.
Besides, fixed receptive fields will limit the network's ability to model objects of different sizes.
To choose a proper receptive field that can be adaptive to the distribution of voxels, a receptive field selection block is designed to choose a
suitable window size for each voxel. 

Features of each location need to interact with each other or be aggregated to learn the local or global context.
There are three typical ways to achieve this: 1) average pooling; 2) convolution; and 3) max pooling.
The former two are linear and parameterized (fixed for average pooling and learnable for convolution); the last one is nonlinear and non-parametric.
Considering the sparsity and non-uniformity of 3D voxels, the shared linear interactive ways would make the feature of a voxel too small (for voxels having few neighbors) or over-smoothed (for voxels having many neighbors). 
Max-pooling, in turn, will always draw the feature of a voxel to the most informative one.
We illustrate the features learned by different interactive manners in Figure~\ref{fig:vis_efs}.
We also compare these three different 
interactive manners
in section~\ref{sec:nonlinearity} and 
the experiments show max pooling works best, which supports our viewpoint that 
more non-linearity is critical for 
3D scene understanding.

The overall LRP module can be 
formulated as
\begin{align}
  S &=(S_1, S_2, S_3) = \text{linear}(x), \\
  P_{1} = p(x, 1), P_{2} &=  p(P_1, 3), P_{3} =  p(P_2, 9),  \\
  Output &= S_{1}P_1  + S_{2}P_2 + S_{3}P_3, 
\label{eq_LRP}
\end{align}
where $x$ denotes the input features, 
S is the selected weight, which is divided into $S_1, S_2, S_3$ for selecting the features of different window sizes,
and $p(x, k)$ means max pooling with kernel size 3 and dilation k.
Here, k is set as 1, 3 and 9 by default.

\subsection{Long Range Pooling Network}
\label{sec:network}
\begin{figure}[t]
    \centering
    \includegraphics[width=\columnwidth]{ 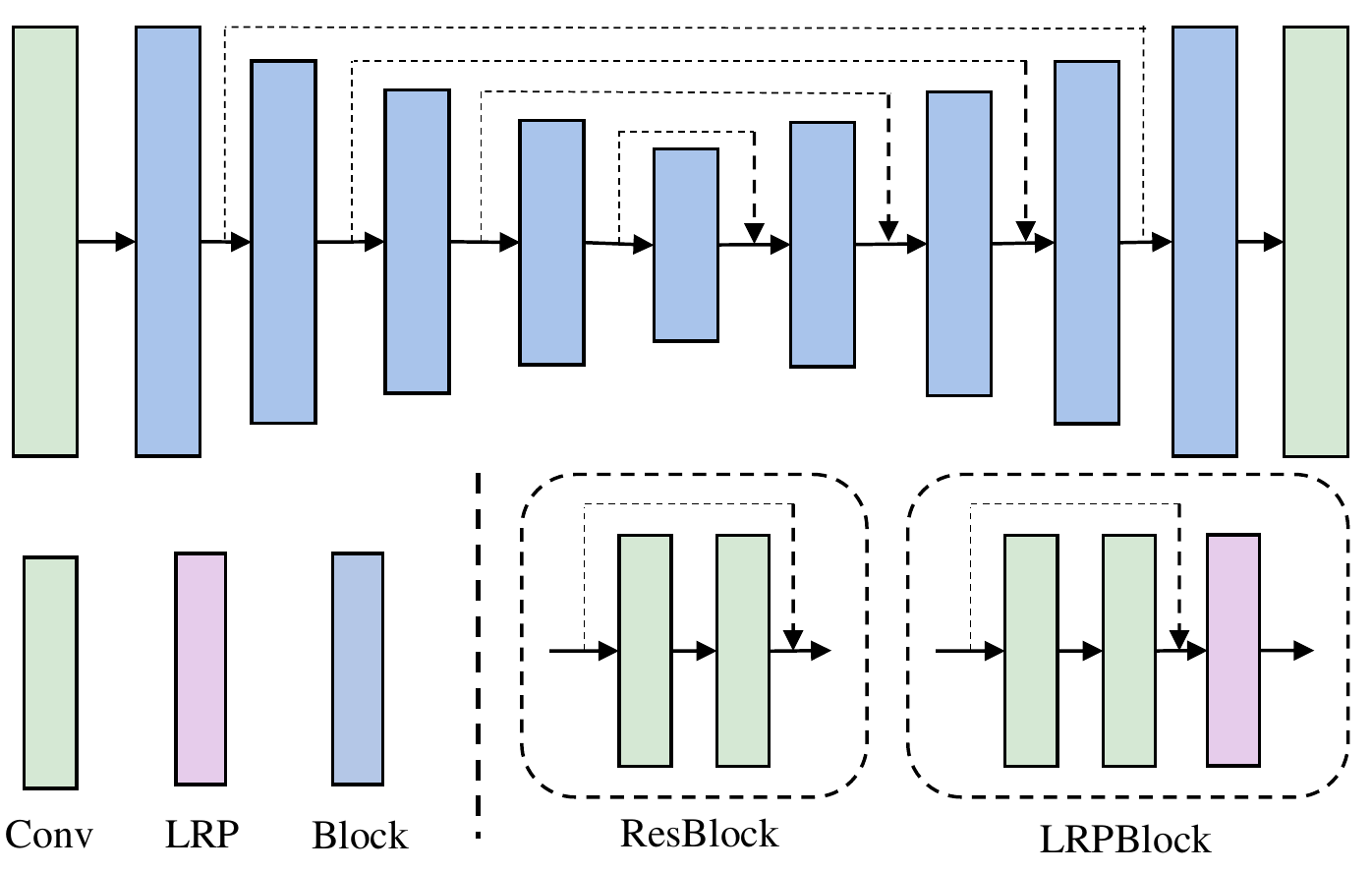}
    \caption{Network architecture for scene segmentation. We use the sparse convolutional U-Net~\cite{hu2021vmnet} as our baseline, consisting of four stages. For the baseline, we use ResBlock as the basic block to extract features, and we replace the ResBlock with LRPBlock to build LRPNet. }
    \label{fig:network}
\end{figure}

In order to verify the effectiveness of LRP module for processing 3D scene data, we have designed LRPNet, based on the U-Net architecture commonly used in previous work~\cite{hu2021vmnet,schult2020dualconvmesh,choy20194d,chen2022scaling}. 
We use the sparse convolutional U-Net implemented by VMNet~\cite{hu2021vmnet} as the baseline. 
Figure~\ref{fig:network} shows the baseline model, a U-Net network with 4 stages; the basic module is a ResBlock,  built from two 3D Convolution+BatchNorm+ReLU blocks and an additional residual path.  

As we have already noted, the LRP module can readily be incorporated into existing networks. Therefore, we build LRPNet by simply adding an LRP module after ResBlock in the baseline model.
In addition to placing the LRP behind the ResBlock, we alternatively experimented with placing it front, middle, and as a parallel branch. These experiments and analysis are discussed in Section~\ref{sec:lrp_position}.

\section{Experiments}
We performed various experiments on several semantic segmentation datasets, to verify our claims and compare our results to those of other methods.

\subsection{Datasets and metrics}
ScanNet v2~\cite{dai2017scannet} is a large-scene benchmark, with 1,613 scenes split 1201:312:100  for training, validation and testing. It has 20 semantic classes. We follow the same protocol as previous works~\cite{choy20194d, graham20183d} and use mean class intersection over union (mIoU) as the metric to evaluate results. Since the annotations of the test set on ScanNetv2~\cite{dai2017scannet} are not available, we conducted the runtime complexity experiments and ablation 
experiments on the validation set.

S3DIS~\cite{armeni20163d} is a large-scene semantic parsing benchmark,  containing 6 areas with 271 rooms,  annotated with 13 categories. Following~\cite{choy20194d,thomas2019kpconv,wu2022point,guo2021pct}, we use Area 5 as the test set, other areas for training and mIoU for evaluation. We also report overall point-wise accuracy (OA) and mean class accuracy (mAcc).

Matterport3D~\cite{chang2017matterport3d} is a large-scene RGB-D dataset, with 90 building-scale scenes annotated in 21 categories.
Since the semantic labels are annotated for faces, we project them to the vertices using the same method as in ~\cite{schult2020dualconvmesh}, and then test on scene vertices directly.
Following~\cite{schult2020dualconvmesh,hu2021vmnet}, we use mAcc for evaluation.

\subsection{Implementation details}
Our experiments were all conducted on the full scenes without cropping. During training and inferencing, we just used vertex colors as input. Following~\cite{hu2021vmnet,choy20194d}, we voxelized the input point cloud at a resolution of 2 cm for ScanNet v2. Following common practice, the input points of S3DIS and Matterport3D are voxelized at a resolution of 5 cm.
During inferencing, we calculated metrics by projecting the predictions back to the raw point clouds using the nearest neighbor. 

During training, we performed scaling, z-axis rotation, translation and chromatic jitter for data augmentation~\cite{hu2021vmnet}. 
For all datasets, we minimized the cross entropy loss using the SGD optimizer with a poly scheduler decaying from a learning rate of 0.1, following~\cite{hu2021vmnet,choy20194d,chen2022scaling}.
Following previous work~\cite{schult2020dualconvmesh,huang2019texturenet}, we used class weights for cross entropy loss on Matterport3D. For ScanNet v2 and Matterport3D, we trained our model for 500 epochs with batch size set to 8, and for S3DIS, we trained for 1,000 epochs with batch size set to 4.

Our experiments were conducted on RTX 3090 GPUs and A5000 GPUs. All runtime complexity experiments are conducted on one RTX 3090 GPU.

\begin{table}
\caption{mIoU (\%) scores for various methods on the ScanNet v2 3D semantic benchmark, for validation and test sets. The best number is in boldface. ``-'' means the number is unavailable.}
\label{tab:scannet_benchmark}
\begin{center}
\begin{tabular}{l|r|rrr}
  \toprule
  Method & Input & Val & Test\\ 
  \midrule
    PointNet++~\cite{qi2017pointnet++} & point & 53.5 & 55.7 \\
    3DMV [26] & point& - &48.4\\
    PointCNN~\cite{li2018pointcnn} &point& - &45.8\\
    PointConv~\cite{wu2019pointconv} &point &61.0 &66.6\\
    JointPointBased~\cite{chiang2019unified} &point& 69.2 &63.4\\
    PointASNL~\cite{yan2020pointasnl} &point &63.5 &66.6\\
    RandLA-Net~\cite{hu2020randla} &point & - & 64.5\\
    KPConv~\cite{thomas2019kpconv} & point & 69.2 & 68.6\\
    PointTransformer~\cite{zhao2021point} & point & 70.6 & - \\
    SparseConvNet~\cite{graham20183d} & voxel & 69.3 &72.5\\
    MinkowskiNet~\cite{choy20194d} & voxel & 72.2 & 73.6 \\
    LargeKernel3D~\cite{chen2022scaling} & voxel & 73.2 & 73.9 \\
    Fast Point Transformer~\cite{park2022fast} & voxel & 72.1 & - \\
    Stratified Transformer~\cite{lai2022stratified} & point & 74.3 & 73.7 \\
  \midrule
    LRPNet (ours) & voxel & \textbf{75.0} & \textbf{74.2}\\
  \bottomrule
\end{tabular}
\end{center}
\end{table}

\begin{table}
\setlength\tabcolsep{2.5pt}
\caption{Several scores (\%) for various methods on the S3DIS segmentation benchmark. The best number is in boldface. ``-'' means the number is unavailable.}
\label{tab:s3dis_benchmark}
\begin{center}

\begin{tabular}{l|r|rrr}
  \toprule
  Method & Input & OA & mAcc & mIoU\\ 
  \midrule
    PointNet~\cite{qi2017pointnet} & point & - & 49.0 & 41.1 \\
    SegCloud~\cite{tchapmi2017segcloud} & point & - & 57.4 & 48.9 \\
    TangentConv~\cite{tatarchenko2018tangent} & point & - & 62.2 & 52.6 \\
    PointCNN~\cite{li2018pointcnn} & point & 85.9 & 63.9 & 57.3 \\
    HPEIN~\cite{jiang2019hierarchical} & point & 87.2 & 68.3 & 61.9 \\
    GACNet~\cite{wang2019graph} & point & 87.8 & - & 62.9 \\
    PAT~\cite{yang2019modeling} & point & - & 70.8 & 60.1 \\
    ParamConv~\cite{wang2018deep} & point & - & 67.0 & 58.3 \\
    SPGraph~\cite{landrieu2018large} & point & 86.4 & 66.5 & 58.0 \\
    PCT~\cite{guo2021pct} & point & - & 67.7 & 61.3 \\
    SegGCN~\cite{lei2020seggcn} & point & 88.2 & 70.4 & 63.6 \\
    PAConv~\cite{xu2021paconv} & point & - & - & 66.6 \\
    KPConv~\cite{thomas2019kpconv} & point & - & 72.8 & 67.1 \\
    MinkowskiNet~\cite{choy20194d} & voxel & - & 71.7 & 65.4 \\
    Fast Point Transformer~\cite{park2022fast} & voxel & - & 77.3 & 70.1 \\
    PointTransformer~\cite{zhao2021point} & point & 90.8 & 76.5 & 70.4 \\
    Stratified Transformer~\cite{lai2022stratified} & point & \textbf{91.5} & \textbf{78.1} & \textbf{72.0} \\
  \midrule
    LRPNet (ours) & voxel & 90.8 & 74.9 & 69.1 \\
  \bottomrule
\end{tabular}
\end{center}
\end{table}

\begin{table*}[t]%
\setlength\tabcolsep{1.5pt}
\caption{Mean class accuracy (\%) scores on the Matterport3D test set. The best number is in boldface.}
\label{tab:matterport_bechmark}
\resizebox{\textwidth}{!}{
\begin{tabular}{l|c|rrrrrrrrrrrrrrrrrrrrr}
  \toprule
  Method & mAcc & wall & floor & cab & bed &chair &sofa &table &door &wind &shf &pic &cntr &desk &curt &ceil &fridg &show &toil &sink &bath &other\\
  
  \midrule
SplatNet~\cite{su2018splatnet}  & 26.7 & \textbf{90.8} & 95.7 & 30.3 & 19.9 & 77.6 & 36.9 & 19.8 & 33.6 & 15.8 & 15.7 & 0.0 & 0.0 & 0.0 & 12.3 & 75.7 & 0.0 & 0.0 & 10.6 & 4.1 & 20.3 & 1.7 \\
PointNet++~\cite{qi2017pointnet++} & 43.8 & 80.1 & 81.3 & 34.1 & 71.8 & 59.7 & 63.5 & \textbf{58.1} & 49.6 & 28.7 & 1.1 & 34.3 & 10.1 & 0.0 & 68.8 & 79.3 & 0.0 & 29.0 & 70.4 & 29.4 & 62.1 & 8.5\\
ScanComplete~\cite{dai2018scancomplete} & 44.9 & 79.0 & \textbf{95.9 }& 31.9 & 70.4 & 68.7 & 41.4 & 35.1 & 32.0 & 37.5 & 17.5 & 27.0 & 37.2 & 11.8 & 50.4 & \textbf{97.6} & 0.1 & 15.7 & 74.9 & 44.4 & 53.5 & 21.8 \\
TangentConv~\cite{tatarchenko2018tangent} & 46.8 & 56.0 & 87.7 & 41.5 & 73.6 & 60.7 & 69.3 & 38.1 & 55.0 & 30.7 & 33.9 & 50.6 & 38.5 & 19.7 & 48.0 & 45.1 & 22.6 & 35.9 & 50.7 & 49.3 & 56.4 & 16.6 \\
3DMV~\cite{dai20183dmv}  & 56.1 & 79.6 & 95.5 & 59.7 & 82.3 & 70.5 & 73.3 & 48.5 & 64.3 & 55.7 & 8.3 & 55.4 & 34.8 & 2.4 & \textbf{80.1} & 94.8 & 4.7 & 54.0 & 71.1 & 47.5 & 76.7 & 19.9 \\
TextureNet~\cite{huang2019texturenet}  & 63.0 & 63.6 & 91.3 & 47.6 & 82.4 & 66.5 & 64.5 & 45.5 & 69.4 & 60.9 & 30.5 & \textbf{77.0} & \textbf{42.3} & 44.3 & 75.2 & 92.3 & 49.1 & 66.0 & 80.1 & \textbf{60.6} & 86.4 & 27.5\\
DCM-Net~\cite{schult2020dualconvmesh}  & 66.2 & 78.4 & 93.6 & \textbf{64.5} & \textbf{89.5} & 70.0 & \textbf{85.3} & 46.1 & \textbf{81.3} & \textbf{63.4} & 43.7 & 73.2 & 39.9 & 47.9 & 60.3 & 89.3 & 65.8 & 43.7 & 86.0 & 49.6 & 87.5 & \textbf{31.1}\\
VMNet~\cite{hu2021vmnet} & 67.2 & 85.9 & 94.4 & 56.2 & \textbf{89.5} & \textbf{83.7} & 70.0 & 54.0 & 76.7 & 63.2 & 44.6 & 72.1 & 29.1 & 38.4 & 79.7 & 94.5 & 47.6 & 80.1 & 85.0 & 49.2 & 88.0 & 29.0\\
\midrule
LRPNet (Ours) & \textbf{70.7} & 83.7 & 95.0 & 58.0 & 88.2 & 81.3 & 79.0 & 54.3 & 78.5 & 60.0 & \textbf{63.4} & 70.7 & \textbf{48.7} &\textbf{ 52.0} & 70.0 & 93.7 & \textbf{66.1} & \textbf{87.4} & \textbf{89.0} & 47.2 & \textbf{88.1} & 30.5\\
  \bottomrule
\end{tabular}}
\end{table*}
\subsection{Comparison to other methods}

We compared our model to other point-based and voxel-based state-of-the-art methods on ScanNet v2, S3DIS and Matterport3D, with results shown in Tables~\ref{tab:scannet_benchmark}--\ref{tab:matterport_bechmark}, respectively.
For these datasets, our method gave the best overall results compared CNN-based methods.
In particular, our method surpasses both the popular state-of-the-art MinkowskiNet~\cite{choy20194d}, and the voxel and mesh fusion algorithm VMNet~\cite{hu2021vmnet}.
Our method even outperforms transformer-based methods, which learn the long range context at a cost of heavy computation, on ScanNetv2.

In addition, we also compared the number of network parameters used and the speed of our algorithm to two SOTA voxel methods~\cite{choy20194d,graham20183d} and two SOTA transformer-based methods~\cite{park2022fast,lai2022stratified} using the ScanNet v2 validation set.
For a fair comparison, we tested the validation set with 312 scenes and average the inference time of each scene to calculate the runtime complexity. We used the MinkowskiNet~\cite{choy20194d} implemented by Fast Point Transformer~\cite{park2022fast}, which is faster than the original implementation.
Table~\ref{tab:complexity_benchmark} shows that our algorithm achieves better results with fewer parameters and faster speed. 
Notice that our method is nearly $16 \times$ faster than Stratified Transformer~\cite{lai2022stratified}.
LRPNet achieved a significant increase in mIoU by just adding the LRP module to the end of each stage of the baseline, with only a few extra parameters introduced by the selection module.

\begin{table}
\setlength\tabcolsep{1.5pt}
\begin{center}
\caption{Number of network parameters and speed for various models on the ScanNet v2 validation set.}
\label{tab:complexity_benchmark}
\resizebox{\columnwidth}{!}{
\begin{tabular}{lrrr}
  \toprule
  Method & Params (M) & Runtime (ms) & mIoU (\%)\\ 
  \midrule
    SparseConvNet~\cite{graham20183d} & 30.1 & 173.5 &69.3\\
    MinkowskiNet~\cite{choy20194d} & 37.9 & 166.1 & 72.2 \\
    Fast Point Transformer~\cite{park2022fast} & 37.9 & 341.4 & 72.0 \\
    Stratified Transformer~\cite{lai2022stratified} & 18.8 & 1149.9 & 74.3 \\
  \midrule
    Baseline & 8.1 & 38.1 & 71.2 \\
    LRPNet (Ours) & 8.5 & 67.9 & 75.0\\
  \bottomrule
\end{tabular}}
\end{center}
\end{table}

\begin{figure*}[t]
    \centering
    \includegraphics[width=\textwidth]{ 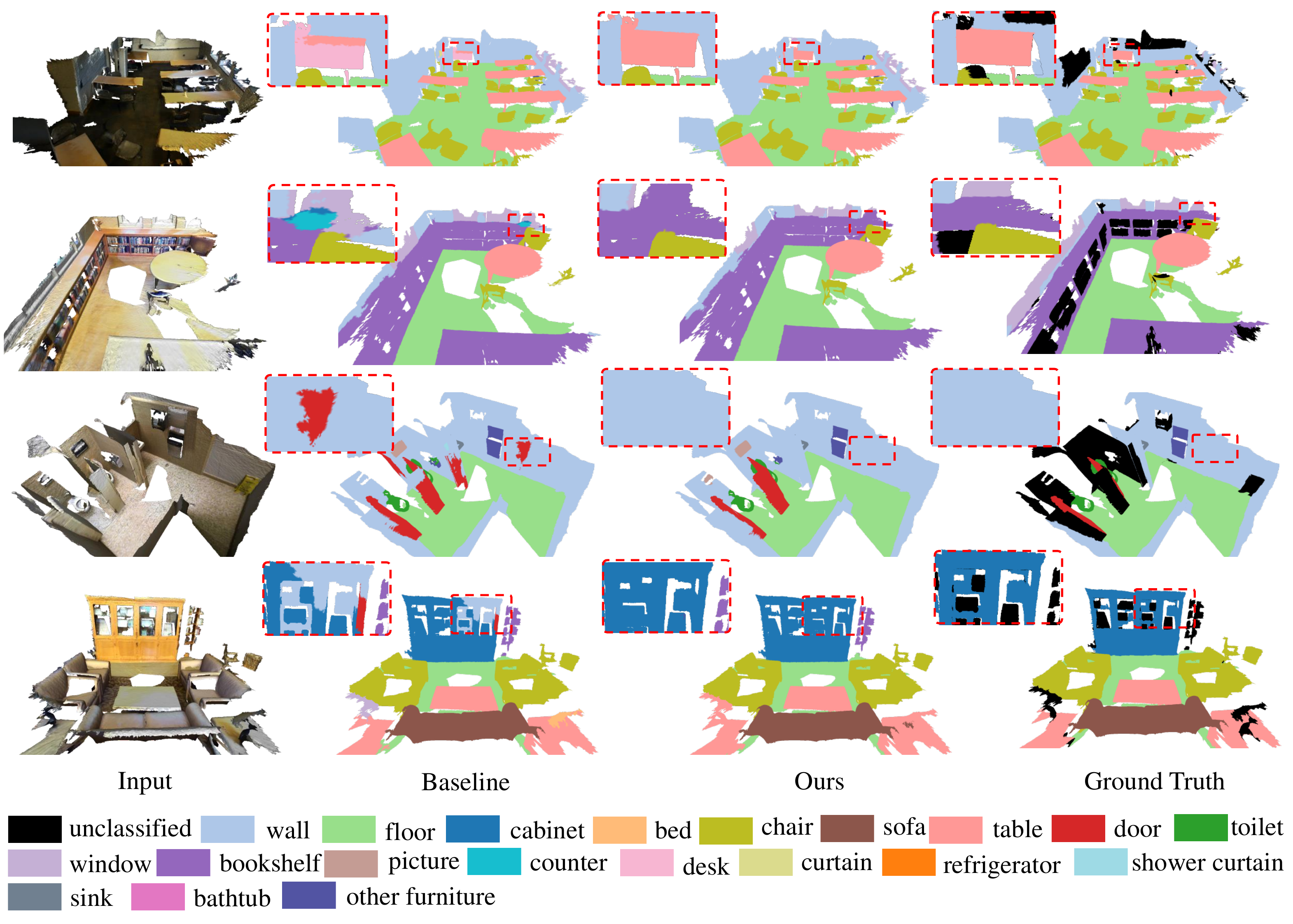}
    \caption{Various ScanNet v2 validation set segmentation results. Red dotted boxes highlight differences between our results and the baseline results.}
    \label{fig:scannet}
\end{figure*}

We further compared qualitative results on the ScanNet v2 validation set, for our model and the baseline model. Figure~\ref{fig:scannet} shows  LRPNet's segmentations are closer to the ground truth than the baseline's segmentations. Our method still works well even in hard cases, e.g.\ for the table and the cabinet in the corner in the first and four rows, whereas the baseline method does not.  LRPNet is also more capable of segmenting out broken objects, as in the second and third rows of bookshelves and walls in Figure~\ref{fig:scannet}.

\begin{table}
\setlength\tabcolsep{1.5pt}
\caption{Ablation study on LRP module. \textbf{Baseline}: U-Net described in section~\ref{sec:network}. \textbf{MaxPool}: The max pooling used in LRP. \textbf{Dilation}: Dilation used in max pooling (default for LRP is 1, 3, 9). \textbf{Selection}: the selection module in LRP. \textbf{Params}: parameters of the network. \textbf{Runtime}: the average time for inferencing one scene. \textbf{mIoU}: the segmentation accuracy metric (\%).}
\label{tab:lrp_ablation}
\begin{center}
\resizebox{\columnwidth}{!}{
\begin{tabular}{cccc|rrr}
  \toprule
  Baseline & MaxPool & Dilation & Selection & Params (M) & Runtime (ms) & mIoU\\ 
  \midrule
   \checkmark  &    &   &  & 8.1 & 38.1 & 71.2 \\
   \checkmark  & \checkmark & & & 8.1 & 74.2 & 72.6 \\
   \checkmark  & \checkmark & \checkmark &  & 8.1 & 65.0 & 73.7 \\
   \checkmark  & \checkmark & & \checkmark & 8.5 & 76.4 & 73.1\\
   \checkmark  & \checkmark & \checkmark & \checkmark & 8.5 & 67.9 & 75.0 \\
  \bottomrule
\end{tabular}}
\end{center}
\end{table}

\subsection{Ablation study}
\label{sec:modeldesign}
In section~\ref{sec:lrp}, we use dilation max pooling to expand the effective receptive field, and use a selection module to allow each voxel to select receptive fields based on features. 
we conducted the ablation experiments on the validation set of ScanNet v2 to verify the design of the LRP module.
Specifically, we choose the sparse convolutional
U-Net implemented by VMNet~\cite{hu2021vmnet} as the baseline, and design new network using max pooling without any dilation. This new network is further enhanced with dilation or/and the receptive field selection module.

As shown in Table~\ref{tab:lrp_ablation}, max pooling can enlarge the receptive field to improve the results, and dilation max pooling can achieve a larger receptive field and a better result. Besides, the selection module can enhance the capability of the network with only introducing a few parameters.

\subsubsection{Position of LRP}
\label{sec:lrp_position}
In section~\ref{sec:network}, 
we added the LRP module \textbf{after} each stage of the baseline (VMNet~\cite{hu2021vmnet}), 
which we name it (\textbf{After}).
Actually, 
we may also put the LRP module before the stage (\textbf{Before}),
in the middle of the stage (\textbf{Middle}), or as an additional path parallel to the stage (\textbf{Parallel}). 
Table~\ref{tab:position_ablation} shows 
the ablation of different settings.
It shows the LRP module 
works better when added
after each stage than in other positions,
also being more computationally friendly. 

\begin{table}
\caption{LRP position study. \textbf{Before}: LRP added before the stages of baseline. \textbf{After}: LRP added after the stages. \textbf{Middle}: LRP added at the middle of stages. \textbf{Parallel}: LRP as an additional branch parallel to the stage.}
\label{tab:position_ablation}
\begin{center}
\begin{tabular}{l|rrr}
  \toprule
  Position  & Params (M) & Runtime (ms) & mIoU (\%)\\ 
  \midrule
   Before  & 8.8 & 77.1 &  74.0 \\
   Middle  & 8.5 & 69.6 & 73.4 \\
   Parallel & 8.5 & 68.4 & 73.1 \\
   After (Ours)   & 8.5 & 67.9 &75.0 \\
  \bottomrule
\end{tabular}
\end{center}
\end{table}

\subsubsection{Operations with non-linearity}
\label{sec:nonlinearity}
As we mentioned in section~\ref{sec:lrp}, max pooling has more non-linearity than average pooling and convolution, and the operations with non-linearity will enhance the capability of the network. 
To investigate the impacts of non-linearity, we conducted non-linearity experiments by replacing the max pooling (\textbf{MaxPool}) of LRP with convolution (\textbf{Conv}) or average pooling (\textbf{AvgPool}). 
In addition, we tested the effect of different receptive fields with these operations (MaxPool, AvgPool, Conv). In Table~\ref{tab:operation_ablation}, $\times N$ is the size of the receptive field of $N\times N \times N$ and $[\times 3, \times 9 \times 27]$ means the outputs of the LRP module is selected from the features with the receptive field of $3 \times 3 \times 3$, $9 \times 9 \times 9$, and $27 \times 27 \times 27$.

As Table~\ref{tab:operation_ablation} shows, although Conv introduces a large number of parameters, the results of MaxPool can still exceed those of AvgPool and Conv in most cases, which proves that the non-linearity of max pooling will enhance the networks for 3D segmentation.

\begin{figure*}[t]
    \centering
    \includegraphics[width=0.97\textwidth]{ 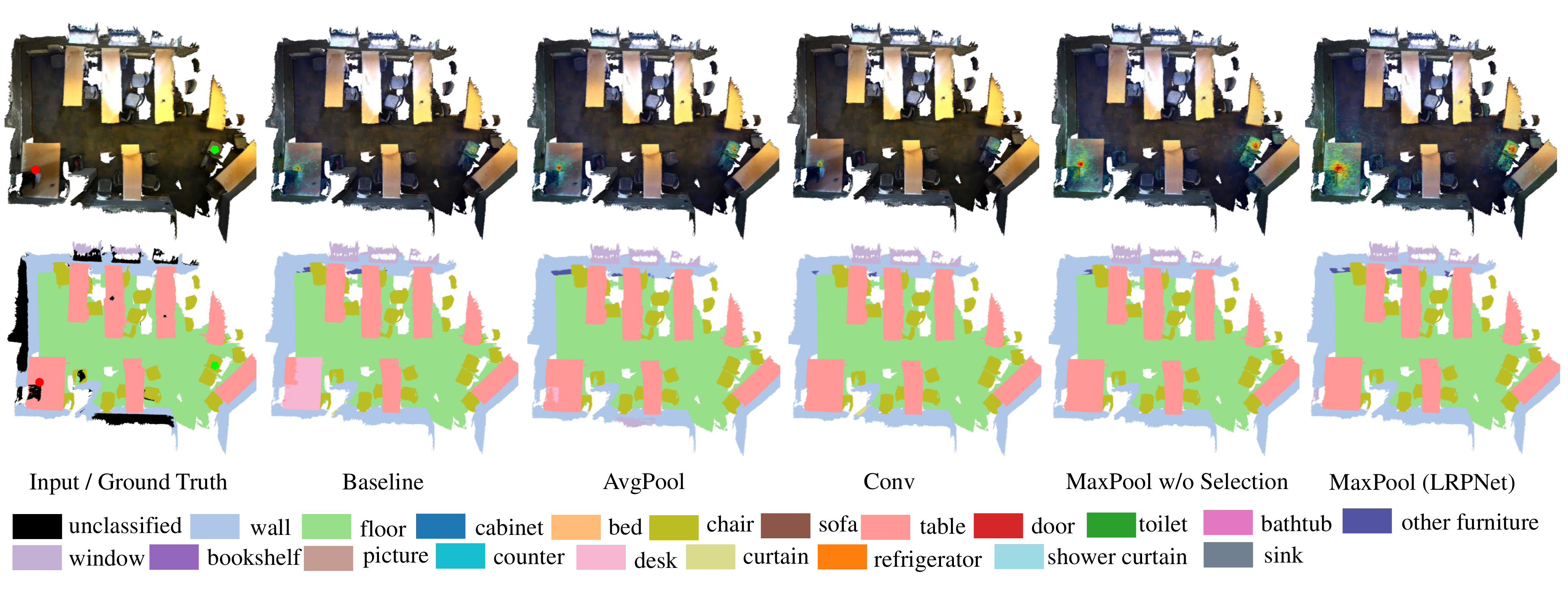}
    \caption{Visualization of Effective Receptive Field (ERF). \textbf{Red Dot and Green Pentagon}: two different positions of interest (marked only in the input). \textbf{First Row}: the ERFs of different methods. \textbf{Second Row}: the ground truth and predictions. \textbf{From Left to Right}: input, baseline, average pooling, convolution, max pooling without selection, max pooling (LRPNet).}
    \label{fig:vis_efs}
\end{figure*}

\begin{table}
\setlength\tabcolsep{2pt}
\caption{Study on non-linearity and range of receptive field. \textbf{MaxPool}: dilation max pooling. \textbf{AvgPool}: dilation average pooling. \textbf{Conv}: dilation convolution. \textbf{Range}: The range and number of receptive fields.}
\label{tab:operation_ablation}
\begin{center}
\resizebox{\columnwidth}{!}{
\begin{tabular}{lr|rrr}
  \toprule
  Method & Range & Params (M) & Runtime (ms) & mIoU (\%)\\ 
  \midrule
   MaxPool  & [$\times$3]  & 8.2 & 66.0 & 72.7 \\
   MaxPool  & [$\times$9]  & 8.2 & 64.8 & 72.8 \\
   MaxPool & [$\times$27]  & 8.2 & 66.1 & 73.9 \\
   MaxPool  & [$\times$9, $\times$27]  & 8.4 & 66.0 & 74.0 \\
   MaxPool & [$\times$3, $\times$9, $\times$27]  & 8.5 & 67.9 & 75.0 \\
   \midrule
   AvgPool  & [ $\times$27] &  8.2 & 58.2 & 73.0\\
   AvgPool  & [$\times$9, $\times$27] & 8.4 & 58.2 &  73.2\\
   AvgPool & [$\times$3, $\times$9, $\times$27] & 8.5 & 60.1 & 73.9\\
   \midrule
   Conv   & [$\times$27] & 17.4 & 65.5 & 72.0\\
   Conv   & [$\times$9, $\times$27] & 17.5 & 66.2 & 73.3\\
   Conv   & [$\times$3, $\times$9, $\times$27] & 17.6 & 66.8  & 73.8\\
  \bottomrule
\end{tabular}}
\end{center}
\end{table}

\subsubsection{Range of LRP module}
As described in section~\ref{sec:method}, while large receptive fields can improve network results, we used dilation max pooling to achieve large receptive fields and a selection module to give the network the ability to select the receptive field size according to the voxel feature.

Table~\ref{tab:operation_ablation} 
shows that as we increase the levels of the receptive field of the LRP module, whether using convolution, max pooling or average pooling, the accuracy of results is gradually improved. We also used different ranges of the receptive fields with the same number of levels. See Table~\ref{tab:operation_ablation} lines 1--3: we fixed the number of receptive fields of LRP module to one level, and then compared results for $[\times3]$, $[\times9]$, and $[\times27]$ receptive fields. With the increasing receptive field for the LRP module, the model also clearly improved.

In addition, we visualized the Effective Receptive Field~\cite{luo2016understanding} of different methods at two positions. As shown in Figure~\ref{fig:vis_efs}, the features of interest are in the center of the table and the center of the chair.
LRPNet has a larger receptive field than baseline, which helps to segment the table better.
Columns 3,4 and 6 in Figure~\ref{fig:vis_efs} show that the operations with linearity can not capture the long range context and their responses are mostly concentrated around the points of interest, while max pooling with non-linearity can make good use of the long range features.
Furthermore, as shown in the last two columns of Figure~\ref{fig:vis_efs}, the selection module can give the network the ability to select the appropriate receptive field according to the voxel features.

\section{Conclusion}
We have proposed a simple yet efficient module, the long range pooling module, 
which can provide an adaptive large  receptive field and 
improve the modeling capacity of the network.
It is easy to add to any existing networks;
we use it to construct LRPNet,
which achieves state-of-the-art results for large-scene segmentation with almost no increase in the number of parameters.
Since the LRP module is simple and easy to use, we hope to apply it to 3D object detection, 3D instance segmentation and 2D image processing, further exploring the principle of improving network performance by use of a large receptive field and operations with greater non-linearity in the future.


{\small
\bibliographystyle{ieee_fullname}
\bibliography{egbib}
}

\end{document}